# The effect of measurement error on clustering algorithms

Paulina Pankowska*[1] and Daniel L. Oberski**

* Vrije Universiteit Amsterdam, The Netherlands

** Utrecht University, The Netherlands

Abstract

Clustering consists of a popular set of techniques used to separate data into interesting groups for further analysis. Many data sources on which clustering is performed are well-known to contain random and systematic measurement errors. Such errors may adversely affect clustering. While several techniques have been developed to deal with this problem, little is known about the effectiveness of these solutions. Moreover, no work to-date has examined the effect of systematic errors on clustering solutions.

In this paper, we perform a Monte Carlo study to investigate the sensitivity of two common clustering algorithms, GMMs with merging and DBSCAN, to random and systematic error. We find that measurement error is particularly problematic when it is systematic and when it affects all variables in the dataset. For the conditions considered here, we also find that the partition-based GMM with merged components is less sensitive to measurement error than the density-based DBSCAN procedure.

---

[1] Corresponding author: Paulina Pankowska, Department of Sociology, Faculty of Social Sciences, Vrije Universiteit Amsterdam, de Boelelaan 1105, 1081 HV Amsterdam, The Netherlands. E-mail: p.k.p.pankowska@vu.nl.

1. Introduction

Clustering is a popular set of statistical techniques widely applied in various scientific disciplines that allows for the separation of data into interesting groups for further analysis or interpretation (Aldenderfer and Blashfield 1984; Dave 1991; Tan, Steinbach, Karpatne, and Kumar 2018). Its main goal is to divide observations, according to their degree of similarity, into a small number of relatively homogenous groups (Bailey 1975). To illustrate, sociologists and economists often use clustering to group career paths and family trajectories, while in psychology and medicine it is commonly applied to identify different variations of an illness or to detect patterns in the spatial or temporal distribution of a disease (McVicar and Anyadike-Danes 2002; Piccarreta and Billari 2007; Tan et al. 2018). In the business world, clustering is performed in the context of customer/market segmentation, a process that divides the market into groups of customers with distinct needs, characteristics, and/or behaviors (Goyat 2011; Tan et al. 2018). In addition, clustering is also frequently used in the fields of pattern recognition, information retrieval, machine learning, and data mining (Tan et al., 2019).

While clustering overall is an important and useful tool (Bailey, 1975), traditional clustering algorithms tend to assume the data are free from measurement error (Kumar and Patel 2007). However, as is well-known, this is an unrealistic assumption. For example, surveys and registers are acknowledged to contain nonnegligible measurement error (Kumar and Patel 2007; Pankowska, Bakker, Oberski, and Pavlopoulos 2018, 2019). In surveys, measurement error is known to result from flaws in the survey response process, the process of data collection, processing, and editing, and from interviewer or respondent effects (Biemer 2004; Sudman, Bradburn, and Schwartz 1997). Errors in register data can be caused by similar factors, but additionally suffer from administrative delay, definition error, and errors caused by administrative incentives (Bakker and Daas 2012; Zhang 2012). Other data sources, such as for instance, weblog data, also contain measurement errors (which are often referred to as "noise") due to the presence of, among other things, online advertisements, navigation panels, copyrights notices, or webpage links from external websites (Onyancha, Plekhanova, and Nelson 2017). All such errors can be considered to have a random (centered i.i.d.) component, as well as a systematic component (location shift and dependence). For example, survey respondents tend to make the same (dependent) errors over time when answering questions (Pankowska et al. 2019).

How do random and systematic measurement error distort conclusions derived from data analysis? For regression and classification, it is well-known how errors bias parameter estimates of interest (see Carroll, Midthune, Freedman, and Kipnis 2006, Fuller 2009, and Gustafson 2003). For example, Pavlopoulos and Vermunt (2015) and Pankowska et al. (2018) demonstrate that estimates of longitudinal turnover in people's employment contracts differ by more than 300 percent —depending on whether measurement error is accounted for or not (estimated turnover proportion decreased from 0.07 to 0.02). However, in the context of clustering, little is known about such effects. On the one hand, errors have the potential to obscure existing clusters, or to produce spurious clusters. On the other, clusters found may still be useful for the purposes at hand – for example interpretation, or relations to external covariates. Indeed, it is difficult to apply the concept of "bias" to the idea of clustering, since this method does not have a universally accepted single purpose (Hennig 2015). In short, while it is clear that data used for clustering have errors, it is not obvious how these errors affect clustering results.

Among the few studies that have investigated the relationship between measurement error and clustering are Dave (1991), which demonstrated the impact of outliers on clustering, and Milligan (1980), which examined the effect of outliers, random error, and nonlinear distortion on clustering. Both concluded that cluster solutions were severely affected, although systematic error was not included in their studies. The effect of systematic error has been investigated in one very specific case, namely in medical diagnostic testing without a gold standard. This field has applied the two-class confirmatory latent class model, in which cluster interpretability is not explored, but assumed (Oberski 2016). In the case in question here, the biasing effects of systematic error on model parameters of interest are well-documented (Hadgu, Dendukuri, and Wang 2012; Torrance-Rynard and Walter 1997; Vacek 1985; Van Smeden Oberski, Reitsma, Vermunt, Moons, and De Groot 2016). However, this work does not extend to more exploratory techniques, which may be focused on interpreting clusters and/or employing them for further analysis.

The observation that errors may affect clustering motivated the development of new techniques, including fuzzy *c*-means clustering (Bezdek, Ehrlich, and Full 1984), noise clustering (Banfield and Raftery 1993; Dave 1991), outlier-robust partition-based clustering (Davé and Krishnapuram 1997; Gallegos and Ritter 2005; García-Escudero, Gordaliza, Matrán, and Mayo-Iscar 2008), noise-robust density-based clustering (Ester, Kriegel, Sander, and Xu 1996), and other "noise-aware" clustering algorithms

(see Aggarwal and Reddy 2013, Ch. 18, for a review). In recent years, the application of (semi-) supervised and unsupervised deep neural networks to noisy data has sparked a literature on general noise-aware learning algorithms (e.g. Goldberger and Ben-Reuven 2016; Malach and Shalev-Shwartz 2017); these methods have been adapted to clustering as well, with a focus on improving classification performance after clustering (see Jindal, Nokleby, Pressel, and Chen (2019) for an overview). Currently, however, we still lack an understanding of (i) the degree to which systematic — i.e. non-i.i.d. and/or uncentered — error affects traditional clustering techniques, and (ii) the degree to which interpretation-oriented purposes of clustering are affected.

In this paper, we perform a Monte Carlo study to investigate the sensitivity of two commonly used clustering algorithms, the Gaussian mixture model (GMM) and DBSCAN, to differing magnitudes and types of random and systematic measurement errors. These techniques were selected because GMMs are a key member of the model-based clustering family (Bouveyron, Celeux, Murphy, and Raftery 2019), and DBSCAN was motivated specifically by the desire to handle noise (Ester et al. 1996), and therefore provides an interesting comparison. Additionally, DBSCAN, unlike GMMs, can handle non- spherical clusters such as "moon-shaped" clusters. We describe how measurement error affects the number of clusters found and the stability of the clusters, two criteria that lie at the basis of cluster interpretation (Hennig 2015). We also evaluate the similarity to clusters obtained in the absence of measurement error, a measure that can be conceived of as similar to that of "bias" in other techniques.

The remainder of the paper is structured as follows: section 2 first provides some background information on clustering techniques in general and on the GMM and DBSCAN algorithms in particular; it then discusses the topic of measurement error and its potential implications for clustering results. Section 3 explains the simulation setup and section 4 discusses the results of the analysis. Finally, section 5 offers some concluding remarks.

2. Background

2.1 Clustering

Cluster analysis is an umbrella term for a variety of algorithms and methods that are used to discover which observations in a dataset are similar and which dissimilar, given a combination of (measured) characteristics (Romensburg 2004). Thus, the aim of clustering is to group cases such that observations belonging to the same cluster are more alike than those belonging to different clusters (Figueiredo Filho, da Rocha, da Silva Júnior, Paranhos, da Silva, and Duarte 2014; Hair, Black, Babin, and Anderson 2014). As clustering can be seen as a classification problem with unobserved outcomes, it is an "unsupervised" learning problem (Bouveyron et al. 2019; Jain 2010). Other applications include the use of clustering to help generate interesting research questions or hypotheses, as well as for strategic decision making in the management field (Romensburg 2004).

There are numerous clustering algorithms available in the literature. Two commonly used approaches are density-based and model-based clustering (Maimon and Rokach 2005). **Model-based clustering** is a probabilistic approach that assumes that the observed data was generated from a mixture of component models, where each of these component models is a probability distribution (Bouveyron et al. 2019). This clustering method requires predefining the number of clusters (Sammut and Webb 2011). On the other hand, **density-based clustering** is a deterministic method that defines clusters in a data space as contiguous regions with high point density. Clusters are separated from each other by regions of low point density and data points lying in these low-density regions may be classified as outliers or noise. In the density-based clustering literature, the mixture models described above are also known as partition-based clustering. Unlike model-based methods, density-based clustering algorithms do not require the number of clusters as an input parameter (Kriegel, Kröger, Sander, and Zimek 2011), nor do they require the clusters to have a parametrically specified, usually convex, shape – they can therefore be seen as "nonparametric" clustering techniques (Kriegel et al. 2011; Maimon and Rokach 2005).

The following two subsections provide an overview of the GMM and DBSCAN algorithms, two highly popular model- and density- based clustering algorithms that we use in our study.

Gaussian mixture models

Gaussian mixture models (GMMs) are among the most commonly used model-based clustering algorithms (Yeung, Fraley, Murua, Raftery, and Ruzzo 2001). They belong to the family of latent variable models and can be defined as a parametric probability density function consisting of a weighted sum of Gaussian component densities (Reynolds 2015). In other words, GMMs assume that data points are generated from a mixture of a finite (predetermined) number, *K, say,* of Gaussian distributions with unknown mean parameters, variance-covariance matrices, and cluster sizes (weights).

GMM seek to estimate a vector of parameters $\boldsymbol{\theta}_k = \{\boldsymbol{\mu}_k, \boldsymbol{\Sigma}_k, w_k\}$ for each of the $K$ $d$-dimensional multivariate Gaussian distributions that correspond to the clusters of interest, $z$. Conditional on the component $z = k$, the observed data vector $\mathbf{x}$ is assumed to follow the multivariate normal distribution,

$$f(x|\boldsymbol{\mu}_k, \boldsymbol{\Sigma}_k) = \frac{1}{(2\pi)^{\frac{D}{2}}|\boldsymbol{\Sigma}_k|^{\frac{1}{2}}} exp\left\{-\frac{1}{2}(x-\boldsymbol{\mu}_k)'\boldsymbol{\Sigma}_k^{-1}(x-\boldsymbol{\mu}_k)\right\} \qquad (1)$$

The marginal density of the observed variables is simply a weighted sum of these $K$ component densities:

$$p(\mathbf{x} \mid \boldsymbol{\theta}) = \Sigma_{k=1}^K w_k f(\mathbf{x} \mid \boldsymbol{\mu}_k, \boldsymbol{\Sigma}_k) \qquad (2)$$

Where $\mathbf{x}$ is a $d$-dimensional vector of continuous data, $w_k$ is a weight parameter for distribution $k$ ($\Sigma_{k=1}^K w_k = 1$), $\mu_k$ is a $d$-length vector of means and $\Sigma_k$ is a $d$ x $d$ variance-covariance matrix. Constraints can be imposed on this variance-covariance matrix. Common choices are to restrict it to a diagonal matrix (spherical components), to set all within-component covariance matrices equal, $\Sigma_k = \Sigma$ (equal shapes), or to specify a reduced-rank decomposition $\Sigma_k = \Lambda\Lambda' + \Psi$ (mixture of factor analysers) (Bouveyron et al. 2019; McLachlan and Peel 2004).

GMM parameters are estimated by fitting a pre-specified number of multivariate normal distributions to the data using the EM algorithm, iterating, for $t = 1,2,...$, between estimating the posterior

$$\hat{p}^{(t)}(z = k \mid \mathbf{x}) = \frac{p(\mathbf{x} \mid \hat{\boldsymbol{\theta}}^{(t-1)}, z = k)}{p(\mathbf{x} \mid \hat{\boldsymbol{\theta}}^{(t-1)})} \qquad (3)$$

and maximizing the expected likelihood

$$\widehat{\boldsymbol{\theta}}^{(t)} = \arg\max_{\boldsymbol{\theta}} \mathbb{E}_{\hat{p}^{(t)}(z=k|\mathbf{x})}[p(z, \mathbf{x} \mid \boldsymbol{\theta})] \tag{4}$$

These two steps are iterated until convergence of $\widehat{\boldsymbol{\theta}}^{(t)}$ or the marginal likelihood (McLachlan and Peel 2004; Reynolds 2015). Note that the posterior estimates $\hat{p}^{(t)}(z = k \mid \mathbf{x})$ produced as a by-product of this procedure form a soft ("fuzzy") classification procedure for the discrete latent components variable, *z*. Direct optimization of the marginal likelihood $p(\mathbf{x} \mid \boldsymbol{\theta})$ is possible as well, although usually avoided for reasons of algorithmic stability. Bayesian solutions to the estimation problem can be found in Frühwirth-Schnatter (2006).

The Gaussian parametric form restricts within-component shapes to "fuzzy" ellipses, whose contours decline exponentially. Fuzziness in clustering has been suggested in the literature to deal with random noise (e.g. Bezdek et al. 1984). In cases where the original clusters are elliptical, one might therefore expect that GMMs should be robust to (Gaussian) random errors. However, even in such ideal cases, systematic errors can easily distort their shape. For example, mean-regressive measurement error will create nonconvex clusters, which cannot be fitted by a single ellipse.

Although a single ellipse (Gaussian component density contour) cannot fit a nonconvex cluster generated by measurement error, such clusters *can* be approximated by merging multiple ellipses. Thus, if one would find multiple elliptical component densities with closely overlapping likelihoods, merging these components into a single cluster should give a picture of the original shape, built up from merged ellipses. This is the basic idea behind *component merging* in Gaussian mixture modelling (Hennig 2010). In this procedure, a distinction is made between the components found by the Gaussian mixture model on the one hand, and the clusters obtained by merging closely overlapping components on the other. Hennig (2010) gives an overview of several component merging techniques.

DBSCAN

DBSCAN – "Density-Based Spatial Clustering of Applications with Noise" – is a nonparametric, deterministic clustering algorithm which groups together points that are close to each other. The algorithm, developed by Ester et al. (1996) requires two hyperparameters:

(i) $\varepsilon$ (Eps)- the maximum distance between two points for them to be considered neighbors;

(ii) *(minPoints)*- the minimum number of neighboring points required to form a so- called dense region.

Using these two hyperparameters, the algorithm identifies the following:

a. **$\varepsilon$ -neighborhood.** The $\varepsilon$ -neighborhood of point $p$ consists of all points $q$ in the dataset $\mathcal{D}$ which are within an $\varepsilon$ distance from $p$, which is determined using a distance function such as the Manhattan Distance or the Euclidean Distance; formally this can be defined by $\{q \in \mathcal{D} \ |\text{dist}(p,q) \leq \varepsilon\}$;

b. *Core object/point.* A core point is one that contains a number of points equal to or greater than *minPoints* in its $\varepsilon$ *-neighborhood*;

c. *Directly density- reachable points.* Point $q$ is defined as directly density-reachable if it is within the $\varepsilon$ *-neighborhood* of $p$, and $p$ is a core point;

d. *Density-reachable points.* Point $q$ is density reachable from point $p$ if for a chain of objects $p_1, \ldots, p_n$, where $p_1 = p$ and $p_n = q$, $p_{i+1}$ is directly density-reachable from $p_i$, given ε and *minPoints*, for $1 \leq i \leq n$. If $q$ is density-reachable for a core point $p$ but is not itself a core point, it is defined as a ***border point***;

e. ***Density connected points.*** Points $p$ and $q$ are density connected if there exists an object $o \in D$ which is a density-reachable point, given ε and *minPoints*, for both $p$ and $q$;

f. ***(Density-based) cluster.*** A cluster $C$ is a non-empty sub-set of $\mathcal{D}$ that satisfies the following conditions:

- $\forall\, p, q$: if $p \in C$ and $q$ is density-reachable from $p$, given ε and *minPoints, then* $q \in C$ (the so called "maximality" requirement)
- $\forall\, p, q \in C: p$ *is density-connected to q*, given ε and *minPoints.*

g. ***Noise.*** The noise cluster contains the set of points in dataset $\mathcal{D}$ that do not belong to any of the clusters $\{C_1, \dots, C_i\}$; noise = $\{p \in \mathcal{D} \;|\forall\, i : p \notin C_i\}$;

Put simply, given the above, the algorithm starts by randomly selecting a core point $p \in \mathcal{D}$ as a seed. It then finds all points in the dataset that are density-reachable from that seed and forms a cluster from a combination of the seed and these points. This process is repeated until all points in the dataset are assigned to a cluster or are classified as noise. DBSCAN is widely used, particularly in the data mining community, due to its flexibility, as it does not require the clusters to be of any specific shape or form (Birant and Kut 2007; Ester et al. 1996).

2.2 Measurement error and its impact on clustering

Measurement error, which is often referred to as "noise" in the data science literature, occurs when the measured or observed value of a variable differs from its true value

(Everitt and Skrondal 2002). Thus, in the context of continuous variables, measurement error can be defined as the difference between the true and measured/observed value of a variable. The error can be either random, i.e. occurring by chance without a specific pattern, or systematic, e.g. such that either consistently under- or overestimates the values of a variable, is dependent on certain characteristics, or is subject to autocorrelation. Overall, measurement error has been shown to severely affect model estimates and lead to biased results (Crocker and Algina 1986; Pankowska et al. 2018, 2019).

Formally, for a given random variable $X$ and its observed counterpart $Y$, e.g. an individual characteristic such as income that is measured using a survey question, measurement error can be conceptualized in the following way:

$$Y = X + \varepsilon \tag{5}$$

Where $\varepsilon$ is the measurement error term and, thus, in the absence of measurement error $Y = X$. When measurement error is random, we can think of $\varepsilon$ as a normally distributed random quantity that is uncorrelated with $X$ and is i.i.d, i.e. $\varepsilon \sim N(0, \sigma)$ and so $E[Y] = E[X]$. This is to say that in the presence of random measurement error, the observed value of random variable $X$ differs from its true value in a way that is uncorrelated with $X$ and which does not exhibit any specific patterns. In the survey context, such error occurs, for instance, when some individuals due to chance only either over- or underreport their income.

Systematic error (also referred to as systematic bias) can occur for a number of reasons. To illustrate, some survey respondents might systematically overreport their income due to social desirability bias (Hariri and Lassen 2017). In this case $\varepsilon$ can be defined as a normally distributed random variable that is independent of $X$ and i.i.d but such that $E[Y] > E[X]$; that is $\varepsilon \sim N(\mu, \sigma)$, where $\mu \neq 0$.

When the probability of making an error depends on a covariate $Z$ that is uncorrelated with $X$, e.g. when the likelihood of overreporting one's income depends on whether the interview was conducted by proxy, we can think of $\varepsilon$ as no longer an i.i.d random variable but rather one whose distribution parameters are some function of $Z$. In other words, while $\varepsilon$ remains independent of $X$ it is only i.i.d conditional on $Z$ and can be defined as follows:

$$\varepsilon \sim \begin{cases} N(\mu_0, \sigma_0) \text{ if } Z = 0 \\ N(\mu_1, \sigma_1) \text{ if } Z = 1 \end{cases}, \text{ where } \mu_1 > \mu_0 \qquad (6)$$

Finally, if the probability of misreporting income depends on the level of income itself, then $\varepsilon$ is both no longer independent of $X$ nor is it i.i.d. In this case, the relationship $\varepsilon \sim N(\mu, \sigma)$ still holds, but is extended in such a way that $\mu$ could be some monotonic function of $X$, with the substantive implication being that higher income individuals are more likely to misreport their income:

$$\varepsilon \sim N(\mu, \sigma), \text{ where } \mu = f(X) \qquad (7)$$

The impact of the aforementioned types of measurement error on clustering specifically has not been studied extensively. Although the overall research on the topic is scarce, the literature available (which concerns solely random types of errors), does argue that clustering algorithms are likely to be (substantially) affected by measurement error (Dave 1991; Frigui and Krishnapuram 1996; Kumar and Patel 2007). One of the few papers actually examining this impact is by Milligan (1980). The author investigates the effects of different types of error perturbation on the results of two types of clustering (hierarchical and k-means) and concludes that in many cases the presence of error in the data leads to a degradation in cluster recovery. This analysis, however, focuses predominantly on random error/noise and does not investigate the impact of systematic errors.

As mentioned above, given the lack of comprehensive evidence regarding the effects of measurement error on clustering, our simulation study looks at how different types and magnitudes of both random and systematic errors affect two aspects of clustering results. More specifically, we look at the number of clusters, as well as the similarity of the clusters to the "original" ones (i.e. those obtained in the absence of error). The choice of the GMM and DBSCAN algorithms (in addition to being driven by their popularity and wide application) is motivated by their potential to mitigate some of the effects of measurement error. More specifically, the attractiveness of GMMs is linked to the fact that they are probabilistic models and so can account for some of the uncertainty introduced by measurement error. The DBSCAN algorithm is used in our analysis as it includes a noise cluster, which might potentially capture (some of the) observations that contain measurement error and leave the substantial clusters, to an extent, intact. The setup of the simulation study is discussed in detail in the next subsection.

3. Simulation setup

As stated above, we use a simulation analysis to demonstrate the effect of different degrees and types of measurement error on DBSCAN and GMM estimates. In more detail, our approach is to first generate a "baseline" dataset containing no measurement error, and then to compare model outcomes on that and error-induced datasets. These steps will be explained in more detail below. Also, as an illustration, the appendix provides pseudocode for generating the "baseline" dataset and introducing measurement error according to one condition.

Step I: Simulating the "baseline" dataset and performing clustering

First, we generated the initial/original, error-free dataset. In essence, our aim was to create a simple dataset consisting of a mixture of multivariate Gaussians, which will ensure strong internal cohesion (homogeneity) and external isolation (separation) of estimated clusters. A more complex data structure could negatively affect cluster recovery and lead to a situation wherein the algorithms produce different results for the same dataset, even in the absence of measurement error, due to random model variability. In this case, it would be difficult if not impossible for us to separate the effect of the data structure from that of introducing measurement error on the clustering results. As such, we drew $n = 1000$ observations from a mixture of three multivariate normal (MVN) distributions, with deterministic proportions of 0.4, 0.35, 0.25. To rephrase, this is to say that the first 400 observations were drawn from $MVN\ A$, the next 350 from $MVN\ B$, and the final 250 were taken from $MVN\ C$. As each MVN had dimensionality of three (corresponding to variables $X_1, X_2$ and $X_3$), the end result was a (1000, 3) matrix of random variables. To ensure the aforementioned separation of sample clusters, we used the following population parameters for our simulation[2]:

---

[2] It is worthwhile noting that these population parameters were selected at random; the only consideration was obtaining spherical, fully separated clusters in the absence of measurement error.

$$G_1 \sim N(\boldsymbol{\mu_1}, \boldsymbol{\Sigma_1}) \text{ where } \boldsymbol{\mu_1} = \begin{bmatrix} -2 \\ 9 \\ 12 \end{bmatrix} \text{ and } \boldsymbol{\Sigma_1} = \begin{bmatrix} 1.50 & 0.30 & 0.20 \\ 0.30 & 0.80 & 0.15 \\ 0.20 & 0.15 & 1.30 \end{bmatrix}; n_1 = 400$$

$$G_2 \sim N(\boldsymbol{\mu_2}, \boldsymbol{\Sigma_2}) \text{ where } \boldsymbol{\mu_2} = \begin{bmatrix} 5 \\ 11 \\ 18 \end{bmatrix} \text{ and } \boldsymbol{\Sigma_2} = \begin{bmatrix} 2.00 & 0.40 & 0.15 \\ 0.40 & 1.60 & 0.25 \\ 0.15 & 0.25 & 1.00 \end{bmatrix}; n_2 = 350$$

$$G_3 \sim N(\boldsymbol{\mu_3}, \boldsymbol{\Sigma_3}) \text{ where } \boldsymbol{\mu_3} = \begin{bmatrix} 4 \\ 4 \\ 5 \end{bmatrix} \text{ and } \boldsymbol{\Sigma_3} = \begin{bmatrix} 1.70 & 0.60 & 0.30 \\ 0.60 & 1.50 & 0.40 \\ 0.30 & 0.40 & 1.45 \end{bmatrix}; n_3 = 250$$

The draws from these multivariate Gaussians (constituting the dataset) correspond to the visualization depicted in Figure 1.

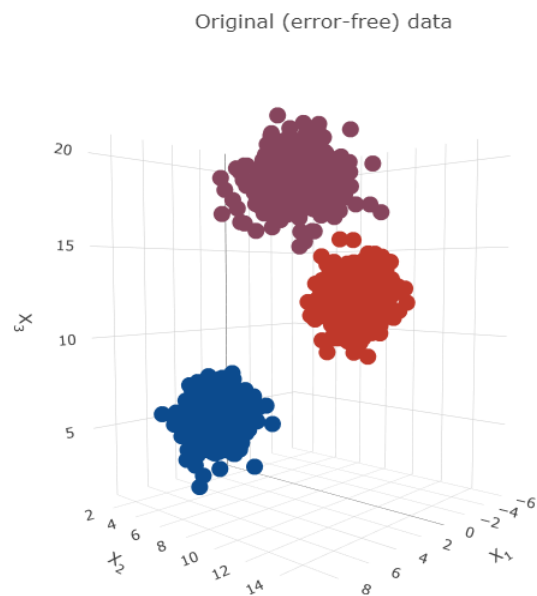

Figure 1- 3D scatterplot of the ""baseline" simulated dataset

Next, we performed clustering on the simulated dataset (i.e. the" baseline" dataset) using a Gaussian mixture model (GMM) and DBSCAN clustering. When fitting GMMs, we fit several models with the number of clusters, $k$, varying from 1 to 10 and chose the model with the best model fit, i.e. the lowest BIC. When using DBSCAN, we set the minimum number of neighboring points to be four for all conditions (as it is recommended that $minPoints = no.of\ dimensions + 1$), while we allowed $\varepsilon$ to vary per condition and chose the appropriate distance based on a visual inspection of the *k*-nearest neighbor distance plot.

Step II: Introducing measurement error into the "baseline" dataset and performing clustering

Having fit the models to the "baseline" dataset, we then introduced various types and levels of measurement error into this data. In doing so, we considered a total of 36 conditions, each bootstrapped 100 times. As illustrated in Figure 2, the following factors were varied in the conditions considered:

- Measurement error rate/ proportion of observations subject to error: 0.1, 0.2 vs. 0.4 (3 levels);

- Number of variables containing measurement error: 1 vs. 3 (i.e. all) (2 levels);

- Type of measurement error: random vs. systematic (2 levels);

- The magnitude of measurement error: low, medium vs. high (3 levels).

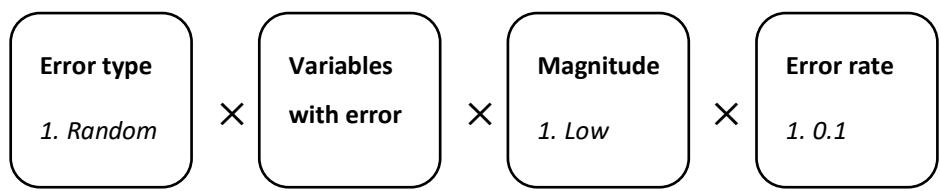

Figure 2- Outline of simulation setup/ simulation conditions

In more detail, for each condition we first randomly selected 0.1, 0.2, or 0.4 of the observations in the dataset. For these observations, we then introduced errors in either one or all three variables. When simulating random error in one of the variables (i.e. only in $X_1$), we added a draw from a normal distribution with $\mu = 0, \sigma =$

{4, 8, 16}. The different $\sigma$'s represent varying degrees of error severity (i.e. low, medium, and high). When introducing systematic error to $X_1$, we added a draw with $\mu = 2.5, 5, 10, \sigma = \{2\}$ wherein the $\mu's$ represent the three different error magnitudes. This is equivalent to the first type of systematic error which is discussed in section 2.2 (i.e. where the error term can be defined as follows: $\varepsilon \sim N\ (\mu, \sigma)$, where $\mu \neq 0$). For the conditions where random error affects all three variables, we added draws normal distributions where:

$$\mu_{x_1} = 0,\ \sigma_{x_1} = \{4, 8, 16\}$$

$$\mu_{x_2} = 0,\ \sigma_{x_2} = \{2, 4, 16\}$$

$$\mu_{x_3} = 0,\ \sigma_{x_3} = \{6, 12, 24\}$$

For systematic error we used the following:

$$\mu_{x_1} = \{2.5, 5, 10\},\ \sigma_{x_1} = \{2\}$$

$$\mu_{x_2} = \{-2.5, -5, -10\},\ \sigma_{x_2} = \{2\}$$

$$\mu_{x_3} = \{1.25, 2.5, 5\},\ \sigma_{x_3} = \{2\}$$

Figure 3 visually shows how the introduction of measurement error affects the simulated dataset, using the following four conditions as illustrative examples: (i) **random** error affecting **one variable** with rate of 0.4 and high magnitude; (ii) **random** error affecting **three variables** with rate of 0.4 and high magnitude; (iii) **systematic** error affecting **one variable** with rate of 0.4 and high magnitude; (iv) **systematic** error affecting **three variables** with rate of 0.4 and high magnitude.[3]

---

[3] The selection of the more extreme conditions (i.e. high error rate and magnitude) was motivated by the fact that their influence on the dataset and subsequently the clustering results is expected to be substantial and highly visible.

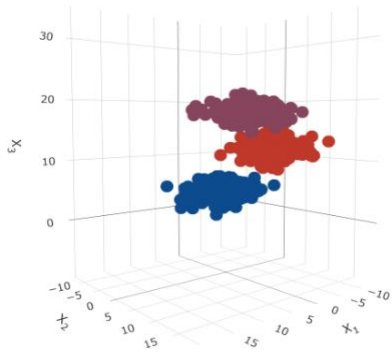 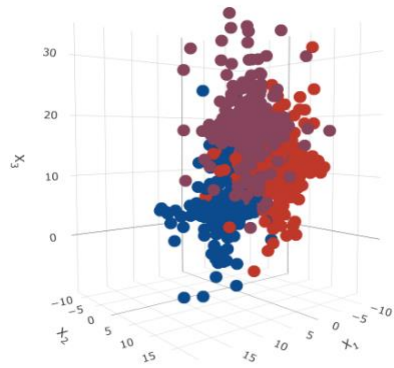
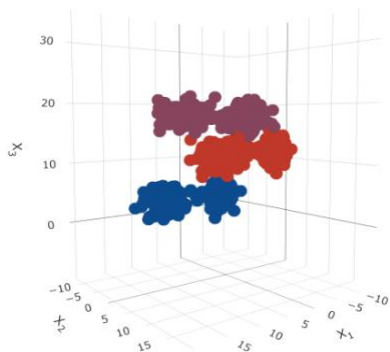 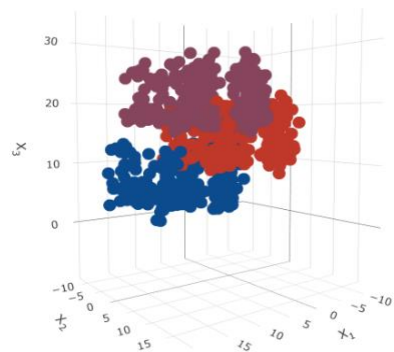

Figure - 3D scatterplot of datasets containing measurement error

As can be seen from Figure 3, when random measurement error affects one of the variables in the dataset, even though it is characterized by high error rate and severe magnitude, the resultant data structure still largely corresponds to the one in the absence of error. That is, even though the groups present in the data appear more "stretched out" and there are some outliers, the data is still characterized by the presence of three, clearly separated groups, which overall seem similar to the original ones. When the random error affects all three variables, on the other hand, the original, three fully separated groups largely overlap; this overlap is likely to impede the recovery of the original clusters. With regards to the condition wherein systematic error affects one variable, as can be deduced from Figure 3, the algorithms are likely to return results that also include spurious clusters. However, as these additional error-driven clusters appear very similar to the original ones, which can still be largely observed in the data, merging the clusters based on similarity (a procedure which is explained in more detail below) might mitigate the impact of this error. Finally, when systematic error affects all three variables, the clusters overlap to such an extent that the clustering algorithms are likely to produce highly dissimilar results to those obtained using the error-free dataset.

Having introduced the error, we then performed clustering on the resultant datasets, using GMM and DBSCAN. In doing so, we followed the same steps as when performing clustering on the "baseline" dataset. Finally, we compared the results to those obtained when no measurement error was introduced (i.e. when using the "baseline" dataset).

Step III: Comparing clustering results in the absence and presence of measurement error

When comparing the results, we focused on two specific metrics: the number of clusters obtained and the similarity of the clusters. While we consider similarity to be of much greater substantive importance, we also look at the number of clusters to understand whether different types of measurement error either obscure clusters or lead to spurious clusters. The importance of the similarity criterion stems from the fact that when the clusters obtained in the presence of measurement error are largely similar to those obtained in its absence, regardless of the number of clusters returned the results can be used for further research or interpretation, and the inferences made should be largely unbiased.

The examination of the number of clusters was relatively straightforward and involved simply comparing the number of the clusters obtained in the absence and presence of measurement error. The evaluation of cluster similarity was carried out based on the adjusted Rand index. The Rand index is a commonly used measure of the similarity between two clusters which varies from 0 to 1, where 0 implies perfect dissimilarity and 1 perfect match (Rand 1971).

The Rand Index can be formalized as follows:

$$\text{Rand Index} = \frac{a + b}{\binom{n}{2}} \tag{8}$$

Where $a$ is the sum of the number of paired observations that are grouped together in the same cluster for both clustering results and $b$ is the number of paired observations that are ungrouped and belong to different clusters for both clustering results. $\binom{n}{2}$ represents the sum of all possible unordered pairs (Rand 1971). The adjusted Rand index is in principle similar to the original index but in addition it accounts for the fact that pairs of observations can be correctly grouped or ungrouped due to chance; it is bounded between ±1 (Hubert and Arabie 1985).

In addition to considering the number and similarity of the clusters obtained directly from the fitted GMM (the so-called mixture components), we also examined the clusters obtained by merging the mixture components. Such merging is a common practice that is applied when the resultant mixture components are not separated sufficiently from one another for them to be interpretable and meaningful. The process is performed in a hierarchical order, whereby the value of a given merging criterion is computed for all pairs of components and the pair with the highest value is merged. Criterion values are then recomputed for the resultant clusters and the merging process continues until the highest criterion value obtained is below a predefined cut-off value. In our application, we use the Bhattacharyya distance as our criterion value and apply a threshold of 0.1. For further details regarding the merging process and the criterion used, we refer to Hennig (2010).

For DBSCAN, we also compare the numbers and similarity of the "baseline" clusters with a subset of the clusters obtained in the presence of error that includes only stable clusters. To calculate stability, we resampled the datasets for each condition using bootstrapping (50 iterations) and compared the clustering results of the bootstrapped samples to those obtained on the original erroneous datasets. In doing so, we used the Jaccard similarity coefficient, which is defined as the size of the intersection of two clusters divided by the size of the union of these clusters. We considered a cluster to be stable if on average, given the 50 bootstraps we run, the Jaccard coefficient was higher than 0.7. For further details regarding the calculations of cluster stability and the criterion used we refer to Hennig (2007).

The analysis was carried out using the R environment for statistical computing (version 3.4.4). When fitting the algorithms to the datasets as well as when merging the GMM components and checking for cluster stability for the DBSCAN results, we used predominantly the *Flexible Procedures for Clustering (fpc)* package (Hennig 2015).

## 4. Results

### 4.1 Clustering in the absence of measurement error

The clustering results obtained in the absence of measurement error (i.e. using the "baseline" dataset) for both GMM and DBSCAN almost perfectly recover the population parameters used to simulate the data. However, as the population parameters were set to ensure the emergence of three distinct, perfectly separated clusters, this was to be expected. More specifically, for GMM, the model that fits the data best (i.e. has the lowest BIC) correctly classifies all 1,000 observations in the dataset and returns the following three clusters (which are extremely similar to the Gaussian distributions used to simulate the data):

$$G_1 \sim N(\widehat{\boldsymbol{\mu}_1}, \widehat{\boldsymbol{\Sigma}_1}) \text{ where } \widehat{\boldsymbol{\mu}_1} = \begin{bmatrix} -1.93 \\ 8.99 \\ 11.99 \end{bmatrix} \text{ and } \widehat{\boldsymbol{\Sigma}_1} = \begin{bmatrix} 1.63 & 0.36 & 0.36 \\ 0.36 & 0.99 & 0.14 \\ 0.36 & 0.14 & 1.52 \end{bmatrix}; n_1 = 400$$

$$G_2 \sim N(\widehat{\boldsymbol{\mu}_2}, \widehat{\boldsymbol{\Sigma}_2}) \text{ where } \widehat{\boldsymbol{\mu}_2} = \begin{bmatrix} 4.95 \\ 10.98 \\ 17.99 \end{bmatrix} \text{ and } \widehat{\boldsymbol{\Sigma}_2} = \begin{bmatrix} 1.78 & 0.32 & 0.22 \\ 0.32 & 1.41 & 0.26 \\ 0.22 & 0.26 & 0.95 \end{bmatrix}; n_2 = 350$$

$$G_3 \sim N(\widehat{\boldsymbol{\mu}_3}, \widehat{\boldsymbol{\Sigma}_3}) \text{ where } \widehat{\boldsymbol{\mu}_3} = \begin{bmatrix} 4.12 \\ 4.01 \\ 5.06 \end{bmatrix} \text{ and } \widehat{\boldsymbol{\Sigma}_3} = \begin{bmatrix} 1.21 & 0.47 & 0.26 \\ 0.47 & 1.40 & 0.28 \\ 0.26 & 0.28 & 1.53 \end{bmatrix}; n_3 = 250$$

The DBSCAN results return four clusters: three substantive ones and a noise cluster. The centroids of the substantive clusters (calculated based on cluster membership) are as follows:

$$C_1 = \begin{bmatrix} -1.91 \\ 9.02 \\ 12.01 \end{bmatrix}; n_1 = 391$$

$$C_2 = \begin{bmatrix} 4.91 \\ 11.02 \\ 18.00 \end{bmatrix}; n_2 = 331$$

$$C_3 = \begin{bmatrix} 4.10 \\ 3.98 \\ 5.00 \end{bmatrix}; n_3 = 242$$

The algorithm assigns 36 observations to the *noise cluster*, all remaining observations (96.4 percent) are classified correctly.

4.2 Clustering in the presence of measurement error

*GMM estimates*

The results obtained when fitting the GMM algorithm to the datasets containing measurement error (for all 36 conditions) are displayed in Table 1 and Figures 4 and 5. As can be seen, overall, the number of clusters as well as cluster similarity (calculated based on the Adjusted Rand Index) remain largely unaffected by random measurement error, provided that only one of the variables in the dataset is subject to error and that magnitude of this error is relatively low or of medium magnitude. The effect of the error rate appears negligible in this case. When error severity is high, on the other hand, we can observe the emergence of spurious clusters, although cluster similarity remains relatively high. Random measurement error also leads to spurious clusters when it affects all three variables, regardless of its magnitude and the error rate. The similarity between the clusters obtained for these conditions and the "original" ones is inversely related to the error rate and its magnitude (i.e. as the error rate and/or magnitude increase, the similarity between the aforementioned clusters decreases).

The effect of systematic error on the clustering results appears significantly more severe. Namely, virtually all 18 conditions can be characterized by the emergence of spurious clusters. What is more, the similarity measure is only truly high when the error affects one variable and is low in magnitude (regardless of the error rate). The remaining conditions return clusters that are substantially different from those in the "baseline" dataset.

When merging the obtained mixture components into more meaningful clusters, a highly optimistic picture regarding the robustness of GMMs to random measurement error emerges. More specifically, the number of clusters appears largely unaffected by measurement error with the exception of two rather extreme conditions, i.e. when the error affects all three variables, its rate is 0.4, and it is either medium or large in magnitude. In the case of these two scenarios measurement error obscures clusters. The resultant clusters are also in most cases highly similar to those obtained in the absence of error. Again, the two above specified conditions are an exception and lead to the emergence of dissimilar clusters. The clusters obtained under the condition wherein a high in magnitude random error affects all three variables and 0.2 of the observations are also dissimilar to the "original" ones, albeit to a lesser extent.

While merging also improves the clustering results for datasets that contain systematic error, it does so to a lesser extent. That is, systematic error distorts the number of clusters for half of the conditions considered, i.e. when the error affects one variable and is large in magnitude or when it affects all three variables and its severity is either medium or high. Likewise, cluster similarity can also be considered dissatisfactory for these conditions. It is worth mentioning that the adjusted Rand index is particularly low when three variables are subject to medium systematic error and 40 percent of observations are affected, or when the error is large and 20 or 40 percent of the cases are affected.

Table 1- GMM clustering results by simulation condition

| Error type | Var's incl. error | Magnitude | Error rate | No. of clusters | Adj. R Index | Np. of merged clusters | Adj. R Index of merged clusters |
|---|---|---|---|---|---|---|---|
| Random | one | low | 0.1 | 3.0 | 0.999 | 3.0 | 0.999 |
| | | | 0.2 | 3.0 | 0.997 | 3.0 | 0.999 |
| | | | 0.4 | 3.0 | 0.999 | 3.0 | 0.999 |
| | | medium | 0.1 | 3.2 | 0.986 | 3.0 | 0.998 |
| | | | 0.2 | 3.5 | 0.952 | 3.0 | 0.997 |
| | | | 0.4 | 3.2 | 0.979 | 3.0 | 0.998 |
| | | high | 0.1 | 4.5 | 0.914 | 3.1 | 0.993 |
| | | | 0.2 | 5.9 | 0.775 | 3.1 | 0.995 |
| | | | 0.4 | 5.1 | 0.725 | 3.0 | 0.996 |
| | three | low | 0.1 | 4.5 | 0.908 | 3.0 | 0.991 |
| | | | 0.2 | 5.1 | 0.817 | 3.0 | 0.984 |
| | | | 0.4 | 4.8 | 0.739 | 3.0 | 0.965 |
| | | medium | 0.1 | 4.2 | 0.892 | 3.1 | 0.973 |
| | | | 0.2 | 5.8 | 0.742 | 3.2 | 0.952 |
| | | | 0.4 | 5.8 | 0.600 | 2.1 | 0.514 |
| | | high | 0.1 | 4.1 | 0.839 | 3.5 | 0.937 |
| | | | 0.2 | 6.3 | 0.632 | 3.7 | 0.777 |
| | | | 0.4 | 7.1 | 0.460 | 1.8 | 0.314 |
| Systematic | one | low | 0.1 | 3.7 | 0.950 | 3.0 | 0.999 |
| | | | 0.2 | 3.8 | 0.941 | 3.0 | 0.999 |
| | | | 0.4 | 3.1 | 0.992 | 3.0 | 0.999 |
| | | medium | 0.1 | 7.1 | 0.752 | 3.4 | 0.984 |
| | | | 0.2 | 7.5 | 0.643 | 3.2 | 0.987 |
| | | | 0.4 | 6.7 | 0.573 | 3.0 | 0.996 |
| | | high | 0.1 | 6.8 | 0.794 | 4.8 | 0.834 |
| | | | 0.2 | 7.0 | 0.655 | 5.9 | 0.733 |
| | | | 0.4 | 6.0 | 0.586 | 6.0 | 0.587 |
| | three | low | 0.1 | 5.6 | 0.800 | 3.1 | 0.995 |
| | | | 0.2 | 6.3 | 0.685 | 3.0 | 0.991 |
| | | | 0.4 | 5.6 | 0.683 | 3.0 | 0.983 |
| | | medium | 0.1 | 7.8 | 0.654 | 5.2 | 0.888 |
| | | | 0.2 | 8.8 | 0.517 | 4.4 | 0.848 |
| | | | 0.4 | 8.5 | 0.406 | 2.1 | 0.475 |
| | | high | 0.1 | 8.7 | 0.599 | 6.6 | 0.678 |
| | | | 0.2 | 9.0 | 0.380 | 5.2 | 0.373 |
| | | | 0.4 | 8.9 | 0.274 | 3.4 | 0.204 |

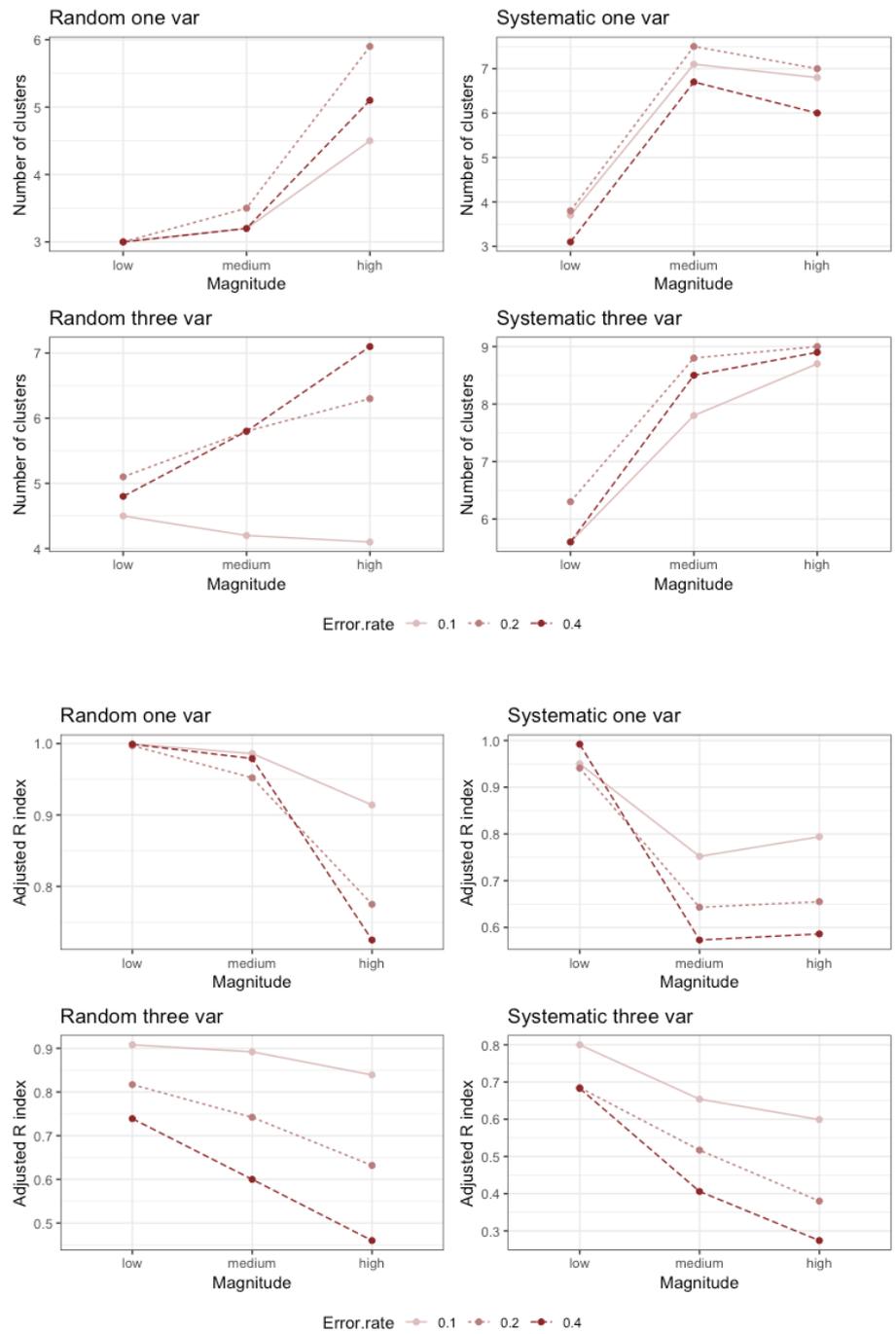

Figure 1- (top) Number of clusters and (bottom) cluster similarity for GMM mixture components

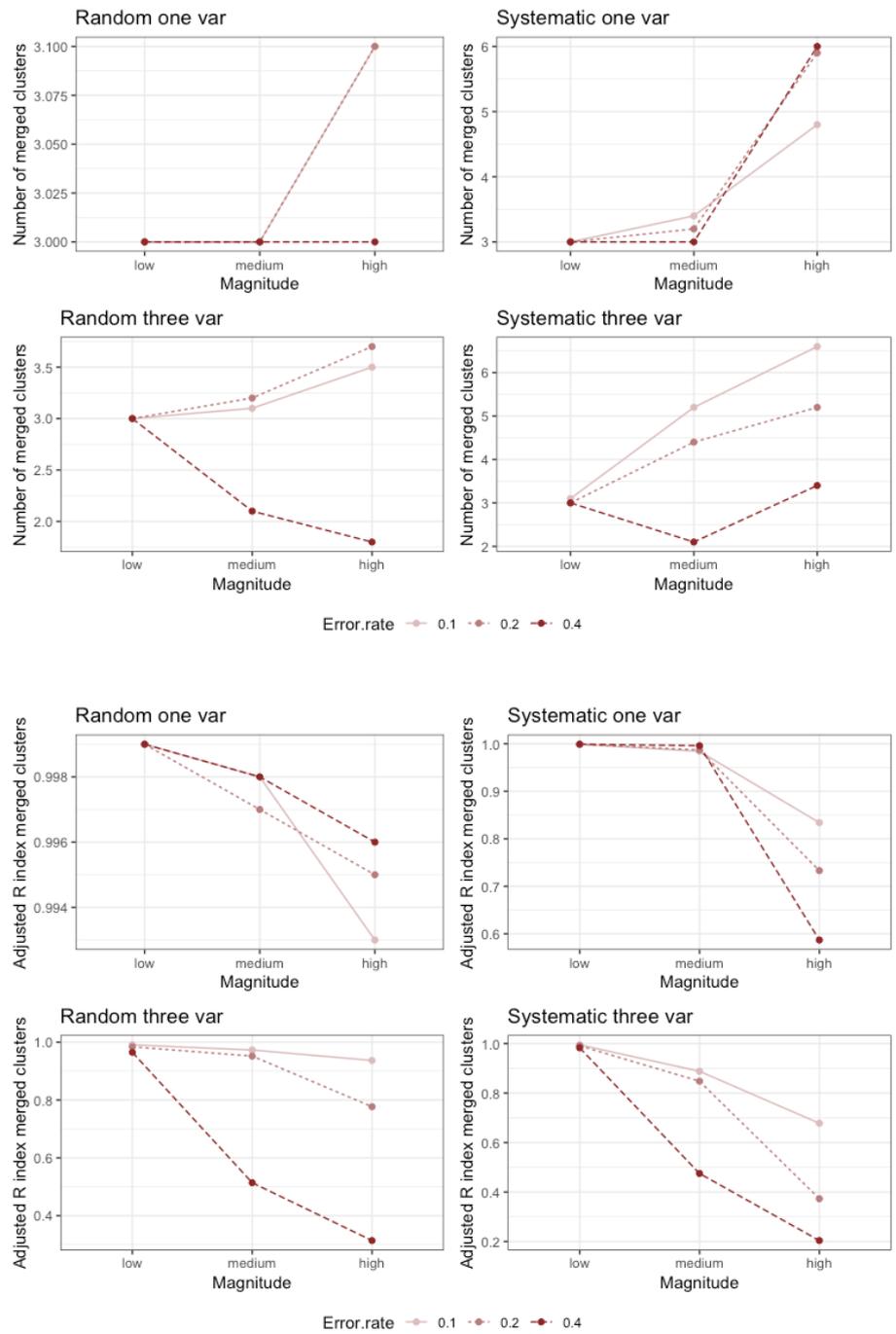

Figure 2- (top) Number of clusters and (bottom) cluster similarity for GMM merged clusters

*DBSCAN estimates*

The DBSCAN results presented in Table 2 and Figure 6 suggest that this clustering algorithm performs worse than GMM in the presence of measurement error. This is particularly striking when looking at cluster similarity. In more detail, when looking at the mean number of clusters obtained in each condition, it can be observed that for most conditions the presence of measurement error does not lead to (many) spurious clusters nor does it obscure clusters. The number of clusters is strongly inflated primarily when the error is systematic and high in magnitude (regardless of the number of variables affected and the error rate).

The degree of similarity of the clusters, however, appears more sensitive to measurement error. That is, DBSCAN returns substantially dissimilar clusters when it is applied to datasets containing random or systematic errors that affect all three variables (with the exception of the conditions in which the error rate is 0.1 and the magnitude low). In other words, for the aforementioned conditions the resultant clusters have very little in common with the ones obtained using the error-free data. Overall, the results do not substantially differ when considering only stable clusters.[4]

Furthermore, as can be seen in Figure 7, contrary to expectations, the noise cluster does not appear to capture observations that are subject to measurement error. This is the case even when the error is random and of very high magnitude, i.e. when the error is anticipated to lead to outliers, which should theoretically be assigned to the noise cluster. More specifically, Figure 7 provides an overview of the size cluster for each of the 36 simulation conditions. For most conditions, the number of observations included in the noise cluster is only slightly higher than the number of observations included in that cluster in the absence of measurement error. More specifically, in most cases the noise cluster size does not exceed 50, while for the error-free data this cluster consists of 36 observations. Therefore, it can be concluded that most observations that were subject to measurement error (i.e. a total of 200 or 400, depending on the condition) were not classified as noise.

---

[4] It is worthwhile mentioning that for all 36 conditions the noise cluster was unstable for most bootstraps.

Table 2- DBSCAN clustering results by simulation condition

| Error type | Var's incl. error | Magnitude | Error rate | No. of clusters | Adj. R Index | No. of stable clusters | Adj. R Index of stable clusters | Size of noise cluster |
|---|---|---|---|---|---|---|---|---|
| Random | one | low | 0.1 | 4.3 | 0.965 | 3.0 | 0.992 | 44.0 |
| | | | 0.2 | 4.2 | 0.946 | 3.1 | 0.971 | 33.0 |
| | | | 0.4 | 4.3 | 0.934 | 3.1 | 0.966 | 35.7 |
| | | medium | 0.1 | 4.3 | 0.946 | 3.2 | 0.975 | 41.6 |
| | | | 0.2 | 4.6 | 0.924 | 3.2 | 0.957 | 35.7 |
| | | | 0.4 | 4.7 | 0.889 | 3.0 | 0.896 | 34.0 |
| | | high | 0.1 | 4.5 | 0.922 | 3.6 | 0.942 | 44.2 |
| | | | 0.2 | 5.2 | 0.879 | 3.2 | 0.909 | 34.3 |
| | | | 0.4 | 4.8 | 0.685 | 2.3 | 0.537 | 30.8 |
| | three | low | 0.1 | 4.4 | 0.909 | 3.5 | 0.938 | 47.4 |
| | | | 0.2 | 4.0 | 0.623 | 2.2 | 0.514 | 38.4 |
| | | | 0.4 | 3.6 | 0.412 | 2.0 | 0.381 | 40.8 |
| | | medium | 0.1 | 3.7 | 0.574 | 2.8 | 0.560 | 26.8 |
| | | | 0.2 | 3.8 | 0.392 | 2.3 | 0.395 | 35.6 |
| | | | 0.4 | 2.9 | 0.054 | 1.3 | 0.049 | 36.9 |
| | | high | 0.1 | 4.9 | 0.620 | 2.9 | 0.671 | 56.2 |
| | | | 0.2 | 3.7 | 0.148 | 1.8 | 0.149 | 43.5 |
| | | | 0.4 | 2.7 | 0.006 | 1.3 | 0.002 | 38.0 |
| Systematic | one | low | 0.1 | 4.7 | 0.951 | 3.1 | 0.991 | 52.6 |
| | | | 0.2 | 4.4 | 0.944 | 3.0 | 0.962 | 28.7 |
| | | | 0.4 | 4.2 | 0.934 | 3.0 | 0.943 | 29.6 |
| | | medium | 0.1 | 5.8 | 0.871 | 3.0 | 0.879 | 39.7 |
| | | | 0.2 | 4.3 | 0.785 | 2.5 | 0.694 | 31.3 |
| | | | 0.4 | 3.7 | 0.700 | 2.1 | 0.522 | 30.3 |
| | | high | 0.1 | 8.5 | 0.822 | 4.2 | 0.914 | 42.6 |
| | | | 0.2 | 7.5 | 0.699 | 5.4 | 0.736 | 31.8 |
| | | | 0.4 | 7.0 | 0.530 | 5.5 | 0.626 | 29.6 |
| | three | low | 0.1 | 4.8 | 0.893 | 3.2 | 0.934 | 54.0 |
| | | | 0.2 | 3.9 | 0.656 | 2.1 | 0.518 | 33.5 |
| | | | 0.4 | 3.2 | 0.409 | 1.6 | 0.330 | 30.6 |
| | | medium | 0.1 | 4.4 | 0.527 | 1.8 | 0.365 | 40.2 |
| | | | 0.2 | 3.5 | 0.233 | 1.4 | 0.228 | 42.4 |
| | | | 0.4 | 2.6 | 0.015 | 1.0 | 0.000 | 35.1 |
| | | high | 0.1 | 8.8 | 0.306 | 6.2 | 0.410 | 38.0 |
| | | | 0.2 | 9.7 | 0.075 | 5.2 | 0.096 | 34.2 |
| | | | 0.4 | 4.3 | 0.062 | 2.3 | 0.074 | 16.5 |

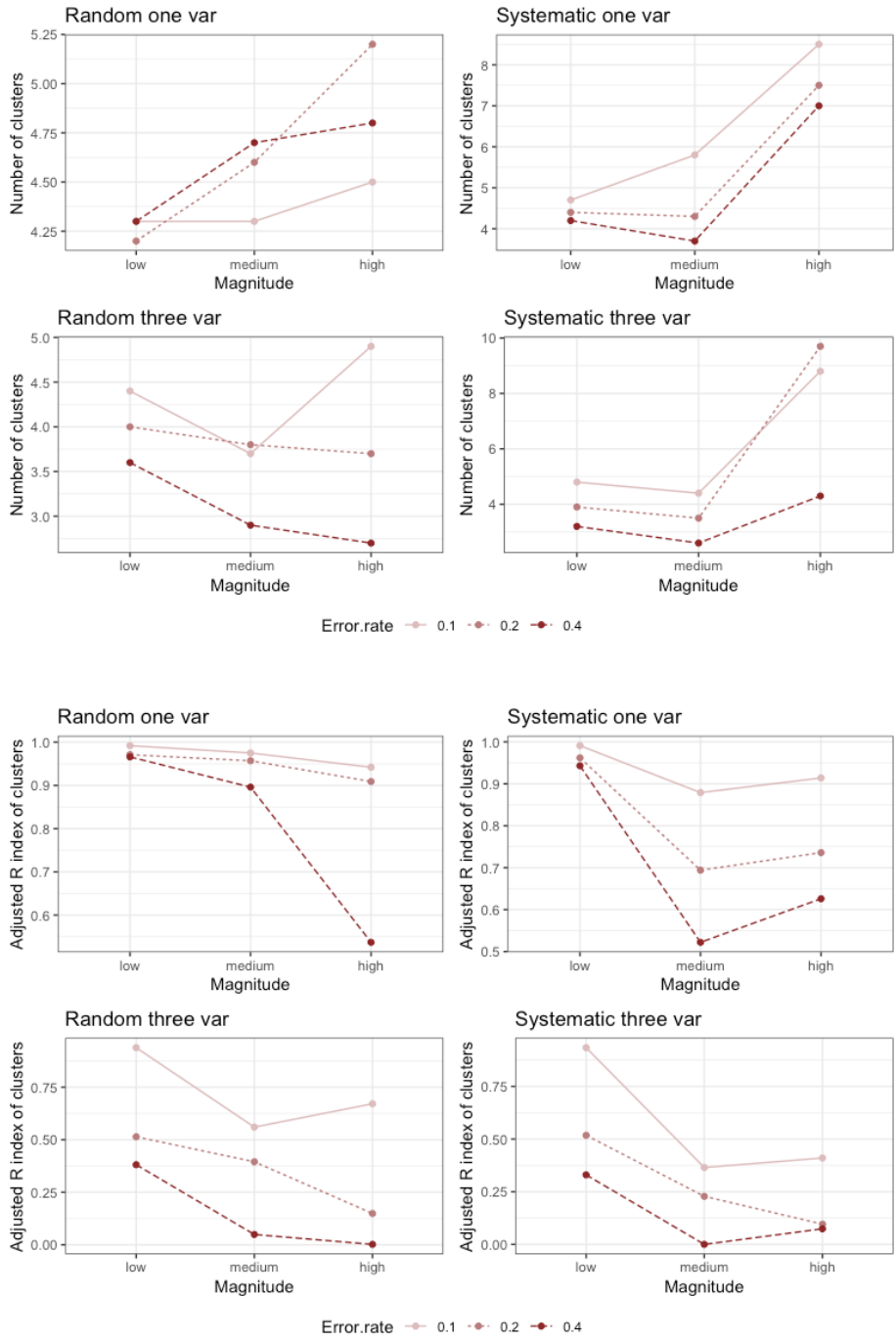

Figure 3- (top) Number of clusters and (bottom) cluster similarity for DBSCAN

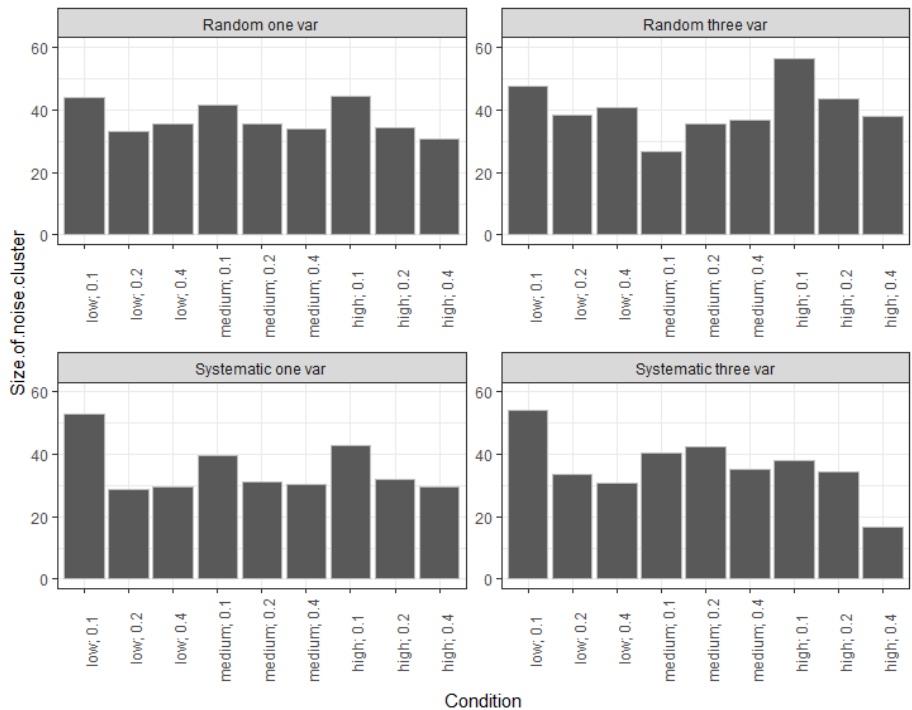

Figure 4- Noise cluster size by simulation condition (where the original noise cluster size = 36)

5. Discussion and conclusions

Clustering is a commonly applied method used in numerous disciplines that allows for the separation of observations into interesting groups, based on a predefined similarity measure. While an important and useful tool, this technique also suffers from an important shortcoming. Namely, in most cases, the clustering algorithms used disregard the problem of measurement error, which is both unrealistic and problematic. It is unrealistic as few, if any, data sources can be truly considered error-free, and it is problematic as measurement error is known to have the potential to severely bias estimates. In the context of clustering, measurement error can, for instance, produce spurious clusters or obscure clusters; it can also affect their shape, form, and stability.

Despite the threat that measurement error poses to the validity of clustering results, research available on the matter is scarce. Therefore, in this paper, we investigated the sensitivity of two commonly used model- and density-based clustering algorithms (i.e. GMMs and DBSCAN) to various types, severities, and levels of measurement error. In doing so, we examined how error affects the number of clusters found, the stability of the clusters, and their similarity to the clusters obtained in the absence of error.

Our results indicate that measurement error is particularly problematic and leads to unreliable clustering results when it is systematic as opposed to random, when it affects all variables rather than only one, and, as expected, when its magnitude and/or rate is high. We also show that, overall, GMM is less sensitive to measurement error than DBSCAN, especially when looking at the merged clusters rather than the mixture components. DBSCAN appears highly sensitive to measurement error, in particular with regards to cluster (dis)similarity, regardless of whether all clusters or only stable clusters are considered. It also appears that, contrary to expectations, the noise cluster of the DBSCAN algorithm does not capture observations with measurement error.

The lower relative sensitivity of GMM estimates to measurement error is a rather surprising result. That is, while GMM can be viewed as the more restrictive clustering algorithm of the two (as, unlike DBSCAN, it makes an explicit assumption about the parametric form of the clusters), it seems to fare better in the presence of measurement error that can distort cluster shapes. These findings, however, should be treated with caution given the data structure of the simulated dataset. More specifically, in our analysis we simulated three almost perfectly spherical clusters and

GMM algorithms are known to perform well when the clusters have such round shapes. DBSCAN, on the other hand, tends to be the preferred clustering method when the shapes of the clusters are arbitrary. Therefore, it is advisable to repeat the analysis using more complex data structures. This will also allow for the investigation of the impact of measurement error in a more realistic setup, as real-world data clusters tend to have various shapes and forms and are rarely perfectly separable.

It is also worthwhile mentioning that, while our analysis focuses on two important and popular types of clustering, it does not investigate the effect of measurement error on hierarchical clustering, a method which is widely used in particular in the social sciences. Therefore, future research should also examine how measurement error impacts such algorithms as Ward's method. We have also only focused on one type of systematic error, i.e. where the values of a variable are systematically over- or underestimated for some randomly selected subset of observations. It would be interesting to also look at how the two other types of systematic error, i.e. errors dependent on covariates or on the true value of the variable itself, affect clustering results.

Finally, given the strong potential implications of measurement error on clustering results, future research should also focus on investigating solutions that allow for the mitigation of its effects. Furthermore, new ways of performing error-aware clustering should consider the diverse nature of measurement error and account for both random and systematic type of errors.

Appendix: Pseudocode illustrating the simulation design

Below we provide an example pseudocode illustrating the simulation design, which corresponds to the condition wherein all three variables contain systematic error that is small in magnitude and that affects 10 percent of the observations. The pseudocode includes the steps taken to simulate the "baseline" dataset and those taken to introduce measurement error according to the condition discussed above.

*Step I: Simulate "baseline" dataset and perform clustering*

1. Draw $n_1 = 400$ observations from the following MVN distribution:

$$G_1 \sim N(\boldsymbol{\mu_1}, \boldsymbol{\Sigma_1}) \text{ where } \boldsymbol{\mu_1} = \begin{bmatrix} -2 \\ 9 \\ 12 \end{bmatrix} \text{ and } \boldsymbol{\Sigma_1} = \begin{bmatrix} 1.50 & 0.30 & 0.20 \\ 0.30 & 0.80 & 0.15 \\ 0.20 & 0.15 & 1.30 \end{bmatrix}$$

2. Draw $n_2 = 400$ observations from the following MVN distribution:

$$G_2 \sim N(\boldsymbol{\mu_2}, \boldsymbol{\Sigma_2}) \text{ where } \boldsymbol{\mu_2} = \begin{bmatrix} 5 \\ 11 \\ 18 \end{bmatrix} \text{ and } \boldsymbol{\Sigma_2} = \begin{bmatrix} 2.00 & 0.40 & 0.15 \\ 0.40 & 1.60 & 0.25 \\ 0.15 & 0.25 & 1.00 \end{bmatrix}$$

3. Draw $n_3 = 400$ observations from the following MVN distribution:

$$G_3 \sim N(\boldsymbol{\mu_3}, \boldsymbol{\Sigma_3}) \text{ where } \boldsymbol{\mu_3} = \begin{bmatrix} 4 \\ 4 \\ 5 \end{bmatrix} \text{ and } \boldsymbol{\Sigma_3} = \begin{bmatrix} 1.70 & 0.60 & 0.30 \\ 0.60 & 1.50 & 0.40 \\ 0.30 & 0.40 & 1.45 \end{bmatrix}$$

4. Perform clustering: fit GMM/ DBSCAN algorithms to the resultant dataset

    a. For GMM: fit models with number of clusters – k – varying from 1 to 10 and chose the model with lowest BIC

    b. For DBSCAN: set the minimum number of neighbouring points to be four; choose the appropriate ε based on the k-nearest neighbour distance plot

*Step II: Introduce measurement error into the "baseline" dataset and perform clustering (100 iterations)*

5. Set the measurement error threshold[5] to 0.1 for all observations (that is $t = 0.1$)

6. For each observation in the dataset, draw a random number from a standard uniform distribution — $U_i \sim U(0,1)$

7. If $U_i \leq t$, add random draws to $X_{1,i}, X_{2,i}$ and $X_{3,i}$ from the following normal distributions $\mu_{x_1} = 2.5 \ and \ \sigma_{x_1} = 2$, $\mu_{x_2} = -2.5 \ and \ \sigma_{x_2} = 2$, and $\mu_{x_3} = 1.25 \ and \ \sigma_{x_3} = 2$

8. Perform clustering using GMM/DBSCAN (as described in (4))

    a. For GMM: also merge mixture components into clusters using a threshold of 0.1 for the Bhattacharyya distance

    b. FOR DBSCAN: also calculate cluster stability using a threshold of 0.7 for the Jaccard coefficient (50 iterations)

*Step III: Compare clustering results in the absence and presence of measurement error*

9. Compare clustering results obtained in (4) and (8)

    a. Compare number of clusters

    b. Compare cluster similarity using the Adjusted Rand Index

---

[5] The threshold corresponds to the probability of being subject to measurement error.